\documentclass[letterpaper]{article}
\usepackage{aaai}
\usepackage{times}
\usepackage{helvet}
\usepackage{courier}

\usepackage{hyperref}       
\usepackage{url}            
\usepackage{booktabs}       
\usepackage{amsfonts}       
\usepackage{nicefrac}       
\usepackage{microtype}      
\usepackage{rotating}
\usepackage{makecell}

\usepackage{amsmath}
\usepackage[]{algorithm2e}
\usepackage{graphics}
\usepackage{graphicx}
\usepackage{caption}
\usepackage{subcaption}

\usepackage{tikz}
\usetikzlibrary{backgrounds, matrix}

\newcommand{\argmin}{\operatornamewithlimits{argmin}}
\newcommand{\argmax}{\operatornamewithlimits{argmax}}

\begin{document}
%
\title{Privacy-preserving Active Learning on \\ Sensitive Data for User Intent Classification}

\author{
  Oluwaseyi Feyisetan \\
  Amazon Research\\
  \texttt{sey@amazon.com} \\
 \And
 Thomas Drake \\
 Amazon Research\\
 \texttt{draket@amazon.com} \\
 \AND
 Borja Balle \\
 Amazon Research\\
 \texttt{pigem@amazon.co.uk} \\
 \And
 Tom Diethe \\
 Amazon Research\\
 \texttt{tdiethe@amazon.co.uk} \\
}

\maketitle
\begin{abstract}
\begin{quote}
Active learning holds promise of significantly reducing data annotation costs while maintaining reasonable model performance. However, it requires sending data to annotators for labeling. This presents a possible privacy leak when the training set includes sensitive user data. In this paper, we describe an approach for carrying out privacy preserving active learning with quantifiable guarantees. We evaluate our approach by showing the tradeoff between privacy, utility and annotation budget on a binary classification task in a active learning setting.
\end{quote}
\end{abstract}

\section{Introduction}
Preserving data privacy is an essential tenet required to maintain the bond of trust between consumers and corporations. Consumers expect their data to remain secure while being used to design better services for them without compromising their identities -- especially while carrying out sensitive transactions and interactions. We define these potentially compromising and personally identifiable data as \textit{sensitive data}. Annotated data drives the machine learning economy and sensitive data holds the key to building richer experiences for users interacting with modern AI interfaces. However, in a bid to get annotations, sensitive data in the wrong hands could lead to irreparable damage in terms of reputation and trust between data holders and their users. 

This potential data transfer deserves greater monitoring in the era of  human powered crowdsourcing and active learning. As niche classification tasks arise to power new applications, they often lack an abundance of pre-annotated datasets. With active learning, the learner can select a subset of the available data points to be annotated. This can exponentially reduce \cite{settles1648active} (in some cases) the number of training queries required. However, the cost \cite{dasgupta2011two,arora2009estimating,settles1648active} of labelling machine learning datasets is traditionally viewed as a function of the expert, time or price.

In this paper, we argue that, for non-public datasets, the cost of learning the true labels should also factor in the privacy of the information contributed by the data owners to the data custodians. As a result, the active learning condition (beyond simply selecting the best examples) becomes two-fold when submitting a \textit{batch} of data for annotation: (1) labeling this selected subset leads to the greatest increase in our machine learning model performance, (2) the probability of revealing any query that can uniquely identify a specific user is very small (and quantifiable by a privacy parameter). 

\paragraph{Contributions} 
Recent studies into privacy and machine learning have focused on preserving model parameters from leaking training data \cite{papernot2016semi,hamm2016learning}. See \cite{ji2014differential} for a recent survey. However, in this paper, we address the privacy preserving requirement from the point of view of training samples that are sent to annotators from an active learning model. To the best of our knowledge, this is the first paper that views preserving privacy in machine learning from this angle. We also describe how techniques such as $k$-anonymity does not provide sufficient privacy guarantees and how this can be improved using differential privacy (DP). We describe how to do this by providing experimental results after discussing an approach that leverages one of the DP algorithms from literature. 

\section{Background}
\label{background}

In this section, we present an introduction to active learning and the privacy challenge of outsourcing queries to the crowd. We then describe $k$-anonymity, its shortcoming in providing an adequate privacy model for active learning and how this can be improved with differential privacy.

\subsection{Active Learning}

The central premise of active learning is that a model is able to perform as well with less data, if a learner can select the training examples that provide the highest information \cite{settles1648active}. Formally described, using a classification task: let $\mathcal{D}$ be a distribution over $\mathcal{X} \times \mathcal{Y}$ where the goal is to output a label $\mathcal{Y}$ from the label space \{$\pm1$\} given an input from the feature space $\mathcal{X}$. The learner receives a \textit{batch} of i.i.d. draws ($x_1,y_1$), ..., ($x_n,y_n$) over the unknown underlying distribution $\mathcal{D}$. The value of $y_i$ is unknown unless an annotation request is made by the learner. The objective is to select a hypothesis function $h: \mathcal{X} \to \mathcal{Y}$, where err($h$) = $\mathbb{P}_{(x,y)\sim\mathcal{D}}[h(x) \neq y]$ is small. Given that $\mathcal{H}$ is the space of all hypothesis, and $h* = \argmin\{$err$(h) : h \in \mathcal{H}\}$ is the hypothesis with minimum error, the aim of active learning is to select a hypothesis $h \in \mathcal{H}$ with error err($h$) within reasonable bounds of err($h*$) by using few annotation requests (i.e., few compared to a passive learner).

Various strategies have been proposed to implement an active learner. One is \textit{uncertainty sampling} \cite{lewis1994sequential} which attempts to select the query $x'$ that the model is least convinced about;  i.e., \begin{math} x' = \argmax_x 1 - P_\theta(\hat{y}|x) \end{math}, where \begin{math}\hat{y}\end{math} is the label with the highest posterior for model \begin{math} \theta \end{math} and $x$ is maximized over the range of all the unlabeled examples in the training pool. Other approaches to uncertainty sampling use either the \textit{margin} between the two most probable classes \begin{math}\hat{y}_1\end{math} and \begin{math}\hat{y}_2\end{math}; i.e., \begin{math} x' = \argmin_x P_\theta(\hat{y}_1|x) - P_\theta(\hat{y}_2|x)\end{math} or a general entropy-based uncertainty over all the possible \begin{math}\hat{y}_i\end{math} classes; i.e., \begin{math} x' = \argmax_x - \sum_i P_\theta(\hat{y}_i|x) log P_\theta(\hat{y}_i|x) \end{math}.

The main privacy issue with active learning stems from the need to scale the annotation process by crowdsourcing the labels via an open call \cite{howe2006rise}. Whenever you make a request to an external resource, you pay a privacy cost by transmitting the information to be annotated. This problem is compounded when there is only one oracle \cite{avidan2007efficient} or collusion among crowd workers. In this paper, we describe privacy notions that can be used to address these concerns along the privacy-utility tradeoff spectrum. 

\subsection{Privacy-preserving machine learning}


\paragraph{\textit{k}-Anonymity} 
At first glance, a straightforward approach for addressing the privacy concerns of active learning could be through $k$-anonymity \cite{sweeney2002k,di2016enforcing}; i.e., ensuring each query that is sent out for crowdsourcing occurs at least $k$ times. In deploying $k$-anonymity, the first step involves identifying a set of \textit{quasi-identifiers}. In our context, these are user queries which can be potentially combined with an externally available dataset to uniquely identify a user. The \textit{frequency set} of these quasi-identifiers represent the number of occurrences in the dataset. We therefore say that a dataset satisfies $k$-anonymity relative to the quasi-identifiers if when it is projected on an external dataset, the frequency set occurs greater than or equal to $k$ times. 


To achieve $k$-anonymity when the size of the frequency set is less than a desired \begin{math} k \end{math}, the attributes are anonymized by either generalizing or suppressing the information. For example, marital status attributes listed as \textit{married, divorced or widowed} are generalized as \textit{once married}, while the ethnicity is redacted as *****. 


Despite its promise, $k$-anonymity has fundamental challenges, some of which are exacerbated by our unstructured data domain. First, \cite{aggarwal2005k} demonstrated that $k$-anonymity suffers from the \textit{curse of dimensionality} since generalization (such as with traditional database columns), requires co-occurrence of words across different examples, but unstructured data such as text phrases tend to follow a heavy-tailed distribution that have a low co-occurrence of words. Secondly, the choice of quasi-identifiers might exclude the selection of some useful sensitive attributes which could then be used for re-identification attacks. This led to other approaches such as $l$-diversity \cite{machanavajjhala2006diversity} and $t$-closeness \cite{li2007t} to handle sensitive attributes. In our implementation, we subsume the quasi-identifiers to include the entire user query.


Therefore, by `hiding in the crowd' of $k$, a user has received some assurance from $k$-anonymity that their sensitive query will not be outsourced from the active learning model unless it passes a meaningful threshold. However, stronger formal privacy guarantees are required to demonstrate that given the user's query, an attacker cannot decide where it came from with certainty. With $k$-anonymity, we are unable to directly quantify a privacy loss value, nor state the bounds of the guarantee of this loss. These two quantities are obtainable from a differential privacy model which we now describe.


\paragraph{Differential Privacy}

To motivate our discourse on why we need stronger privacy guarantees than what $k$-anonymity provides, we consider a hypothetical scenario: Would a user be comfortable asking an AI agent a sensitive question, with the knowledge that the question will be \textit{possibly} used to further train agent's learning model? We denote the training data available to the model before the user submission as $\mathcal{D}$, and the data after the user question as $\mathcal{D}'$. These are \textit{adjacent datasets} differing on only one record. We posit that a user $c$ will be comfortable if (1) $\mathcal{Q(D)} = \mathcal{Q(D')}$ where $\mathcal{Q}$ is a query over the dataset; and (2) $P(\mathcal{S}(c)|D') = P(\mathcal{S}(c))$ where $\mathcal{S}$ is a user secret. These $2$ points are articulated in \textit{Dalenius's Desideratum} \cite{dwork2011firm} that: \begin{quote} Anything that can be learned about a respondent from the statistical database should be learnable without access to the database \end{quote}

\noindent However, we can't make these exact guarantees because datasets are meant to convey information and they will have no utility if these points were true.

What Differential Privacy \cite{dwork2011firm,dwork2014algorithmic} offers is a strong privacy guarantee on adjacent datasets (taking our AI agent example), that: the example selected for active learning will be very similar whether or not the user added their sensitive question. This means, an adversarial annotator receiving a random training query cannot guess with certainty if the query was from dataset $\mathcal{D}$ (which doesn't include the user's query) or $\mathcal{D}'$ (which includes it). 


With this, we state that, a randomized algorithm $\mathcal{M} : \mathbb{N}^{\mathcal{X}} \to \mathcal{C}$ that receives as input a dataset $\mathcal{D}$ with records from a universe $\mathcal{X}$ and outputs an element from $\mathcal{C}$ is $(\varepsilon,\delta)$-differentially private if for every pair of databases $\mathcal{D}$ and $\mathcal{D'}$ differing in one record and every possible set of outputs $C \subseteq \mathcal{C}$ we have
\begin{equation}
\mathsf{Pr}[\mathcal{M}(\mathcal{D}) \in C] \leq e^\varepsilon \mathsf{Pr}[\mathcal{M}(\mathcal{D'}) \in C] + \delta
\end{equation}


The \begin{math} \delta \end{math} parameter accounts for a $  < 1 / ||\mathcal{D}||_1 $ relaxed chance of the \begin{math} \epsilon \end{math} guarantee not holding -- otherwise, it will be equivalent to just selecting a random sample on the order of the size of the dataset. One benefit of the differential privacy model is that it has a quantifiable, non binary value for \textit{privacy loss} which helps in deciding to comparatively select one algorithm over the other. We observe an output of the random algorithm \begin{math} C \sim \mathcal{M}(D) \end{math} where we believe that \begin{math} C \end{math} was more likely produced by $\mathcal{D}$ and not $\mathcal{D'}$, then the privacy loss from the query that yields \begin{math} C \end{math} on an auxiliary input \textit{x} is: \begin{equation} \mathcal{L}(C;\mathcal{M},x,\mathcal{D},\mathcal{D'}) \overset{\underset{\mathrm{def}}{}}{=} \ln(\frac{Pr[\mathcal{M}(\mathcal{D}) = C]}{Pr[\mathcal{M}(\mathcal{D'}) = C]}) \end{equation}

So we surmise that differential privacy promises to prevent a user from sustaining \textit{additional} damage by including their data in a dataset; and the privacy loss obtained is \begin{math} \epsilon \end{math} with probability \begin{math} \ge 1-\delta \end{math}. 

A common method for making the results of a statistical query differentially private involves adding Laplacian noise proportional to either the query's global sensitivity \cite{dwork2008differential,dwork2006calibrating} or the smooth bound of the local sensitivity \cite{nissim2007smooth} (where sensitivity $\Delta f = \max||f\mathcal{D} - f\mathcal{D}'||$). However, for non-continuous domains, adding noise can result in unintended consequences that completely wipe out the utility of the results e.g., \cite{dwork2014algorithmic} describe how attempting to add noise to the query for the optimal price for an auction could drive the revenue to zero. 

Research has however shown that apart from providing reasonable and well understood protection from inadvertent exposure \cite{di2016enforcing},  $k$-anonymity can also be used as a launchpad for achieving quantifiable differential privacy without the utility loss that comes from applying noise \cite{li2012sampling,soria2014enhancing}.

\section{Privacy Preserving Active Learning Framework}
This section introduces our proposed framework for carrying out active learning with privacy guarantees on queries that are sent to an external oracle. It presents the task we try our approach on, highlights the considerations that drive our choices and lays out a high level pseudo-code of our approach.
%
%

\subsection{Task model}
Our task consists of a very large dataset of user queries $\mathcal{U} = \{u_1,...,u_n\}$ that represent the user intent (we map the queries to the intent and do not extract specific quasi-identifiers in order to prevent leakages from un-captured sensitive attributes). Our pipeline consists of an active learning model which learns a binary classifier, predicting if a user intent belongs to a specified class or not. The model is bootstrapped with a golden set of user queries and their associated intents. Subsequent queries from a fixed pool are added to a \textsc{RankedExamplePool} where they are ordered by confidence/uncertainty \cite{gal2016dropout} from our deep learning model. 

To train the model, it first draws on the golden set, then we make a $next\_example$ call to draw an uncertain query from the pool with the criteria that knowing the accurate intent of this query gives the best performance increase to the model while preserving privacy. The query is then outsourced to external annotators and the annotated labels are re-incorporated into the model training process.

\subsection{Considerations} 
Given the size and projected scale of our dataset ($\approx 10^9$ queries), we decide to employ randomized probabilistic algorithms in estimating if a query satisfies $k$-anonymity. Compute and memory resources are thus freed up for training and retraining the model rather than maintaining the frequency and cardinality of incoming queries. Each algorithm (detailed below) is adjusted to prevent over-estimations which could erode the privacy guarantees. Furthermore, after a query is presumed to satisfy $k$-anonymity, only $1$ of the $k$ queries is sent to $n$ external annotators to prevent an aggregation of privacy losses. 

\subsection{Approach}
\label{approach}
In this paper, we adopt the differential privacy algorithm from \cite{li2012sampling} but we utilize it in an active learning setting to select a subset of training examples to send for crowdsourcing. We also note that other DP methods that have been designed for search logs and include a form of $k$ parameter aggregation such as: \cite{korolova2009releasing}, \textsc{Zealous} \cite{gotz2012publishing} and SafeLog \cite{zhang2016safelog} can be implemented to obtain similar results. 

We take a two-stepped approach to extend $k$-anonymity to yield a quantifiable differentially private active learning model taking a cue from how \cite{li2012sampling} demonstrated the use of pre-sampling to achieve differential privacy with $k$-anonymity. This is predicated on \textsc{Theorem 1} from \cite{li2012sampling} which states that: given an algorithm $\mathcal{M}$ which satisfies $(\beta_1, \epsilon_1, \delta_1)$-differential privacy under sampling, then $\mathcal{M}$ also satisfies $(\beta_2, \epsilon_2, \delta_2)$-differential privacy under sampling for any $\beta_2 < \beta_1$ where 
\begin{equation}
\epsilon_2 = \ln \Bigg(1 + \bigg(\frac{\beta_2}{\beta_1}(e^{\epsilon_1} - 1)\bigg)\Bigg); \delta_2 = \frac{\beta_2}{\beta_1}\delta_1
\end{equation}

Therefore, $k$-anonymity on our full dataset (i.e., $\beta_1 = 1$) can instead be preceded by a mechanism that samples each row of its input with probability $\beta_2$, with $k$-anonymity then applied to the resulting sub-sample to yield $\epsilon_2,\delta_2$-differential privacy for $\epsilon_2 = \ln(1 + (\beta_2(e^{\epsilon_1} - 1)))$ within the bounds $\delta_2 = \beta_2\delta_1$. Thus the effect of sampling serves to amplify pre-existing privacy guarantees \cite{balle2018privacy}.


Furthermore, we harden our $k$-anonymity to offer `safe' $k$-anonymization by aggregating the queries by frequency rather than using a distance based measure \cite{lefevre2006mondrian}. The benefit we get from this is that no query within our set of $k$ contains any extraneous sensitive text which could be used as a source of re-identification or to carry out reconstruction attacks.

The next sections describe: how we carry out our sampling to ensure we select useful candidates in an efficient manner, and how we estimate $k$-anonymity using the queries.


\RestyleAlgo{algoruled}
\LinesNumbered
\begin{algorithm}[t]
Let $\beta$ be the sampling rate \tcp{$\beta = \beta_2$ from (3)} 
Let $k$ be the anonymity parameter \\
Let $l$ be number of samples for variance calculation \\
 \KwData{Input multiset of samples $x$ with unknown labels $y$ as: $\mathcal{U} = \{x_1,y_1\},...,\{x_n,y_n\}$}
 \KwResult{$\mathcal{U}'$ filtered private multiset}
\hfill \\
 
Bootstrap \textsc{AL} probablistic model $P_\theta$ with labeled utterances $\mathcal{L}$\\
Retrieve random sub-sample $\mathcal{U}'$ of size $\beta n$\\

\textbf{Pool creation}: to add queries to the \textsc{RankedExamplePool}\\
 \For{$ \{x,y\} \in \mathcal{U}' $}{
  \textbf{retrieve} freq(x); 

  \eIf{freq(x) $\geq$ k}{
   retrieve variance $\phi_x = \textit{var}(P_\theta(\hat{y}|x) : 1 .. l)$ on $u$; \tcp{computed using $l$ draws}
   add $\langle x, \phi_x \rangle$ to \textsc{RankedExamplePool}\;
   }{
   remove $\{x,y\}$ from $\mathcal{U}'$\;
  }
 }
\hfill \\

\textbf{Acquire labels} for top examples sorted by $\phi_x$ descending as $\{\hat{y}_1,...,\hat{y}_{\hat{n}}\}$ \\
Set $\mathcal{U}'$ to be $(x_1,\hat{y}_1),...,(x_{\hat{n}},\hat{y}_{\hat{n}})$ \\
  \textbf{Update model}: draw the next training example from $\mathcal{U}'$ over $\phi_x$\\
\hspace{0.4cm} \textbf{get} $\hat{x} = \argmax_u 1 - P_\theta(\hat{y}|x)$; - \# sample we are least confident of \\
\hspace{0.4cm} retrain learning model using $\{\hat{x},\hat{y}\}$\\

\hfill \\
\textbf{return} $\mathcal{U}'$
 \caption{Privacy-preserving Active Learning}
\end{algorithm}

\section{Efficient sub-sampling for active learning}
\label{estimate_cardinality}

Given a multiset of query sets \begin{math} \mathcal{M} =  \end{math} \{\begin{math}U_1, ..., U_s \end{math}\} with repetitions where a given \begin{math} U_i \end{math} is a tuple \begin{math} \langle u_i, ..., u_k \rangle \end{math}, and a sampling rate $\beta$, our objective is to return a sub-sample from which to carry out $k$-anonymization before training our active learner. Let \textit{n} be the number of distinct query sets |\{\begin{math} U_1, ..., U_s \end{math}\}| with elements \{\begin{math} e_1, ..., e_n \end{math}\}. For a very large dataset size \textit{s}, we seek to estimate \begin{math} \hat{n} \end{math} using only \textit{m} registers where \begin{math} m << n \end{math}. The number of distinct queries in our sample set therefore become $\beta\hat{n}$.

To estimate the cardinality \begin{math} \hat{n} \end{math}, we utlize the \textsc{HyperLogLog} algorithm by \cite{flajolet2007hyperloglog}. \textsc{HyperLogLog} is a probabilistic cardinality estimator that uses a very small memory footprint (\begin{math} \approx \end{math} 12kb per key) for a low standard error (\begin{math} \approx \end{math} $0.81$\%) while scaling up to dataset sizes as large as \begin{math} 2^{64} \end{math} items\footnote{Values taken for the Redis implementation of \textsc{HyperLogLog} - http://antirez.com/news/75}. 

For each incoming \begin{math} U_i : \langle u_i, ..., u_k \rangle \end{math}, a hash \begin{math}h(U_i) \end{math} is computed and converted to base 2. The \textit{b} least significant bits are used to identify the register location to modify, where \begin{math} 2^b = m \end{math} or $\log_2 m$. With the remaining bits \textit{w}, a count \textit{p(w)} is made of the number of running $0s$ up to the leftmost $1$. For a very large, uniformly distributed multiset of random numbers, 2 raised to the maximum value of \textit{p(w)} gives a wide approximate of the cardinality. To correct this, \textsc{HyperLogLog} breaks the multiset into subsets and uses the harmonic mean of the subsets.

After determining our sample size, the next step is to draw a random set of unique samples without replacement up to $\beta\hat{n}$. We keep each element in the dataset with probability $\beta$. The ensuing sub-sample represents the new dataset in our \textsc{RankedExamplePool} from which we will carry out our $k$-anonymization.

\section{Estimating \textit{k}-anonymity using query frequency}
Given a multiset of query sets $ \mathcal{M} =  $ \{$U_1, ..., U_s $\} with repetitions such that the frequency of $U_i$ is $f_{U_i}$ and $ U_i $ is a tuple \begin{math} \langle u_i, ..., u_k \rangle \end{math}. For a very large dataset size $s$, we seek to estimate $\hat{f_{U_i}}$ using sub-linear space. To estimate the query frequency, we use the \textsc{Count-Mean-Min} with conservative update \cite{goyal2012sketch} sketch algorithm which is an improvement on the proposed \textsc{Count-Min} sketch algorithm by \cite{cormode2005improved}. For each incoming $ U_i : \langle q_i, ..., q_k \rangle $, $d$ different hashes of the queries is computed and a counter indexed by each hashed result is incremented. To return the frequency, the minimum over all $d$ index locations for $Q_i$ is returned. To further reduce the potential of error from over-estimation, conservative updates are employed to increment only the minimum counter from the $d$ indexes, and an estimated noise is further deducted from the result.

Therefore after initial pre-sampling step, we select only queries which occur at least $k$ times. These queries are then added to the \textsc{RankedExamplePool} where the \textit{next\_example} is drawn based on the element with the highest uncertainty measure. The benefit of using the frequency to satisfy $k$-anonymity rather than using partitioning, clustering and recoding, or distance based algorithms, is to prevent attacks that rise from an attackers a-priori knowledge of a dataset. For example, a cluster of $k$ with one sensitive or extreme outlier (e.g., a cluster of incomes within zip code with one UHNW outlier becomes easily identifiable by an attacker even though the aggregation was based on nearest neighbors).

\section{Experiments}

Our work seeks to demonstrate quantifiable privacy preserving guarantees in an active learning setting by taking a pre-sampling approach before carrying out $k$-anonymization. We evaluate our approach on an internal dataset used for intent classification on voice devices.

\subsection{Datasets}
The Intent Classifier dataset consists of a subset of queries from February $2018$. The dataset is used to train a model which determines a binary intent for a user. The dataset consists of $2.5$M queries comprising $58$K distinct data points. Each record contains a user query and a label indicating if it is categorized as a \textsc{Positive} or \textsc{Negative} intent query. Part of the dataset has also been previously discussed and described by \cite{yang2018leveraging}. Figures \ref{fig:dataset_summary_3} and \ref{fig:dataset_summary_4} show the nature of the dataset with a histogram and plot of the frequency distribution of the queries. As expected with textual data, there is a long tail of queries which were observed just once (making up $\approx 60\%$ of the dataset). The dataset consists of $63\%$ of queries labelled as \textsc{Positive} intents vs $27\%$ being \textsc{Negative}.

\subsection{Experiment setup}
The experiment task was binary intent classification in an active learning setting. We created a new baseline model which predicts \textsc{Positive} and \textsc{Negative} intents. For the experiments, the model was initially bootstrapped with $1,000$ labeled examples. The active learner then queries a data pool to get a batch of additional training examples to improve the model. The active learning strategy was uncertainty sampling based on confidence scores. 

\begin{figure}
  \begin{subfigure}[b]{0.48\columnwidth}
    \includegraphics[width=\linewidth]{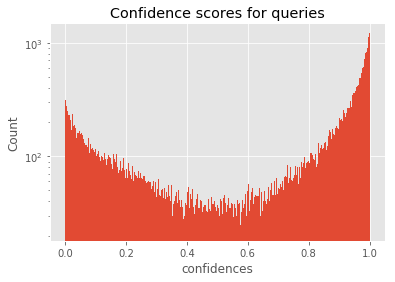}
    \caption{Distribution of confidence scores for each unique query}
    \label{fig:dataset_summary_1}
  \end{subfigure}
  \hfill 
  \begin{subfigure}[b]{0.48\columnwidth}
    \includegraphics[width=\linewidth]{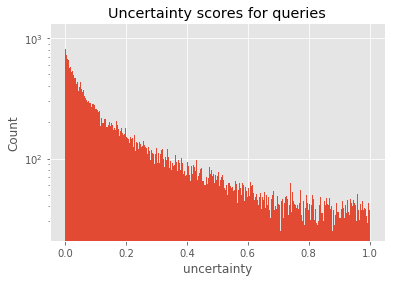}
    \caption{Distribution of uncertainty scores for each unique query}
    \label{fig:dataset_summary_2}
  \end{subfigure}

  \begin{subfigure}[b]{0.48\columnwidth}
    \includegraphics[width=\linewidth]{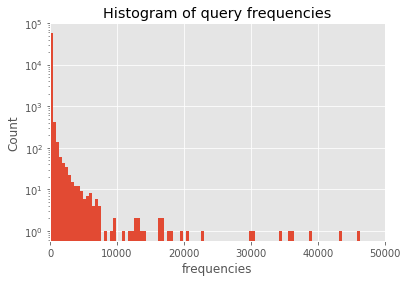}
    \caption{Histogram of frequency distribution}
    \label{fig:dataset_summary_3}
  \end{subfigure}
  \hfill 
  \begin{subfigure}[b]{0.48\columnwidth}
    \includegraphics[width=\linewidth]{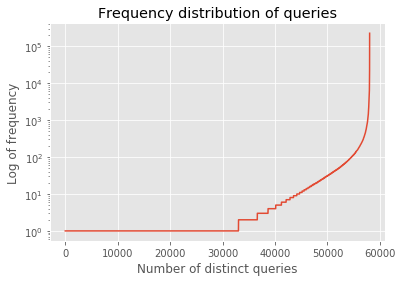}
    \caption{Log frequency distribution of queries}
    \label{fig:dataset_summary_4}
  \end{subfigure}
  \caption{A view into the training dataset}
\label{fig:dataset_summary}
\end{figure}

The confidence and uncertainty scores for the active learning model were obtained from a Bayesian deep learning model described in \cite{yang2018leveraging} where model uncertainty, quantified by Shannon entropy is $\mathcal{U}(x) = -\sum_{c}(\frac{1}{T}\sum_{t} \hat{p}_{c}) \log (\frac{1}{T}\sum_{t} \hat{p}_{c})$ and $\frac{1}{T}\sum_{t} \hat{p}_{c}$ is the averaged predicted probability of class $c$ for $x$, sampled $T$ times by Monte Carlo dropout. A histogram of the confidence and uncertainty scores can be seen in Figures \ref{fig:dataset_summary_1} and \ref{fig:dataset_summary_2}. 

We simulated the probability of the crowd annotators returning the correct answers to the requested queries by drawing from a normal distribution with mean centered at $0.65$ and standard deviation $0.01$ (see \cite{yang2018leveraging}'s Figure $2$(a) for more).


\subsection{Evaluation metrics}
To evaluate our results, we compared the annotation accuracy between the baseline model, and the models trained with active learning and our privacy preserving model. We vary the sub-sampling parameter $\beta$ and the anonymization factor $k$ while training our model and recording its accuracy. We set the evaluation data at $5,000$ samples (i.e., about $10$\% of the dataset). We also provide privacy guarantee values from numerical computations of $\epsilon$ and $\delta$ and highlight in the appendix, what values of $k$ and $\beta$ provide those levels of guarantees.

\paragraph{Baseline condition} train standard classification model. Sub-sampling parameter $\beta = 1$, anonymization factor $k = 1$ i.e., using the entire dataset


\paragraph{Experiment conditions} train classification model using privacy preserving active learning. Sub-sampling parameter varied at $\beta = \{0.1, 0.3, 0.6, 0.9\}$, anonymization factor varied at $k = \{1, 20, 100, 200, 500\}$

\subsection{Results}
The results of our experiments are presented in Figures \ref{fig:accuracy}, \ref{fig:budget} and \ref{fig:budget_accuracy}. Our findings provide insight to the tradeoffs between privacy, utility and our annotation budget. 

\subsection{Privacy vs Utility Tradeoff}
\begin{figure}[h]
  \center{\includegraphics[width=0.8\columnwidth]
  {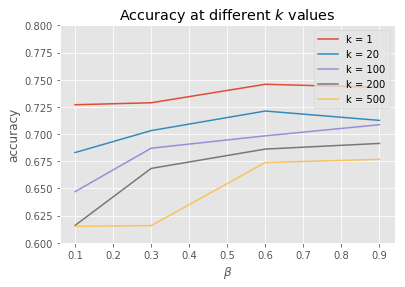}}
  \caption{\label{fig:accuracy} Accuracy at different $k$ values}
\end{figure}

Figure \ref{fig:accuracy} highlights the privacy--utility tradeoff which occurs as a result of varying $\beta$ and $k$. As expected, as the value of $k$, gets smaller, i.e., by selecting more items in the tail of the dataset, we are able to improve the accuracy of our model. This however has the effect of degrading our privacy guarantees. Similarly, by providing privacy amplification by subsampling, the utility of our model suffers. Figure \ref{fig:accuracy} paints a wholistic picture of this by showing how by tuning the values of $\beta$ and $k$, we can arrive at the same values of accuracy.

\subsection{Annotation budget}
\begin{figure}[h]
  \center{\includegraphics[width=0.8\columnwidth]
  {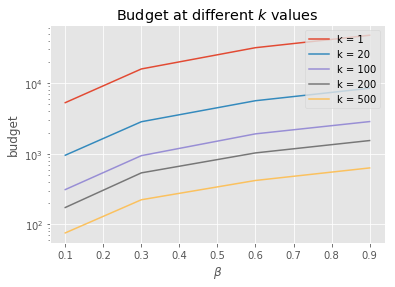}}
  \caption{\label{fig:budget} Budget at different $k$ values}
\end{figure}

Figure \ref{fig:budget} describes how our annotation budget changes for different privacy settings. With a stronger privacy model, we incur less cost as a function of less annotation requests. By reading across the graph, we also discover that the same budget can be realized from different privacy configurations: e.g., by subsampling with $\beta = 0.1$ and selecting $k = 20$, we incur the same budget as $\beta = 0.3$ and $k = 100$ and therefore, the same accuracy (from Figure \ref{fig:accuracy} above). 

\subsection{Budget vs Accuracy}
\begin{figure}[h]
  \center{\includegraphics[width=0.85\columnwidth]
  {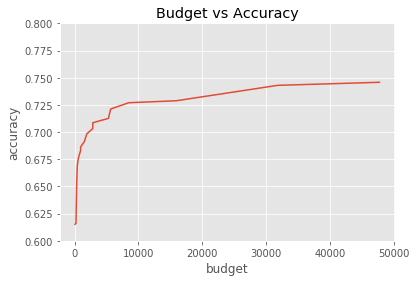}}
  \caption{\label{fig:budget_accuracy} Budget vs Accuracy}
\end{figure}

We established from Figure \ref{fig:accuracy} and Figure \ref{fig:budget}, the relationship between privacy and accuracy, and between privacy and our annotation budget. Since we can obtain the same level of accuracy and budget requirements from different parameter values, Figure \ref{fig:budget_accuracy} highlights how an increase in budget affects our overall model accuracy. Increasing the budget initially accelerates the improvement of our model, however, the utility gains quickly slow down. For example, after $30,000$ labels, we do not see any significant increase in model accuracy.

These results can serve as a guideline in selecting appropriate privacy parameters for different annotation budgets in a way that is more representative of the dataset. For example, for a fixed annotation budget, you can reduce $\beta$ to select more data points from the tail of the dataset (i.e., a smaller $k$). This variation can also be done by starting with a target accuracy score and varying $\beta$ and $k$. The results also demonstrate that by sacrificing some utility gains, we can make stronger privacy guarantees and reduce our annotation budget when carrying out active learning.


\section{Conclusion}
We now briefly revisit our results in the light of our hypothesis. We also discuss the limitations of our process, its implication to the broader discourse on privacy and machine learning and conclude with future work.

We apply the approach from \cite{li2012sampling} to offer privacy guarantees when training models with active learning which requires sending unlabelled examples to an external oracle. Our results join the conversation on differential privacy and machine learning \cite{ji2014differential} with particular reference to preserving the privacy of users.

Our results show that by taking a small performance hit, we can achieve similar accuracy scores with a smaller annotation budget and stronger privacy guarantee. One limitation however is that we have only reported results on a binary classification task. We are currently expanding our approach by designing a new algorithm for differential privacy on text. We show that the accuracy loss increases as task complexity increases. Therefore, if we were to apply the approach in this work to other NLP tasks, e.g. multi-class classification or question answering, we will expect the accuracy loss to be greater.

Another limitation of our approach and potentially other $k-$parameter based approaches (\cite{korolova2009releasing},\cite{gotz2012publishing},\cite{zhang2016safelog}) to differential privacy for text is that it will not work for tasks where almost all the data is unique i.e., $k$ is essentially $1$ (e.g., a datasets of emails or movie reviews). Therefore, a different approach is needed to provide quantifiable privacy guarantees without resorting to $k$-anonymity. 

We believe that this is an area worthy of further research in order to further quantify the true cost of privacy in crowdsourcing and machine learning. We have already begun further work to address two of the limitations reported in this current paper.

\section{Appendix}


Table \ref{epsilon-delta} lays out a grid of $\epsilon$, $\delta$ scores and the corresponding sampling parameter $\beta$ and anonymization factor $k$ required to satisfy that level of ($\epsilon$,$\delta$)-differential privacy. The shaded region presents a $\beta \times k$ high level view of how to achieve a desired level of privacy. 

A few insights can be gleaned from the results, the most obvious being that strong privacy requirements, (indicated by small $\epsilon$ and $\delta$ scores as we traverse the table towards the bottom left corner), require a higher anonymization factor $k$ and smaller sampling rate $\beta$. This is also observed by the fact that lowering the $k$ factor only preserves privacy at the highest displayed $\epsilon$ value of $1.0$ and $\delta = 1 \times 10^{-6}$ (at the top right corner of the table).

Observing $\beta$ individually, we also note that, when $k$ and $\epsilon$ are at fixed values, as $\delta$ decreases, i.e., to obtain stronger privacy guarantees, we need to lower the sampling rate $\beta$. This indicates as $\beta$ decreases, privacy guarantees increase. Similarly, observing $k$ by fixing the values of $\beta$ and $\epsilon$, demonstrates that increasing $k$ improves our privacy guarantees.

\begin{table*}
  \caption{Selected values of $\epsilon$ against $\delta$: the shaded regions show the accuracy scores from our experiments and what $\beta = \{0.1, 0.3, 0.6, 0.9\}$ vs $k = \{20, 100, 200, 500\}$ values provide the ($\epsilon$,$\delta$)-differential privacy guarantee}
  \label{epsilon-delta}
  \centering
  \begin{tabular}{lc|ccccl}
    \toprule
    & & \multicolumn{4}{c}{$\epsilon$}                   \\
    & $\delta$ & \hspace{25pt}0.25 & 0.5 & 0.75 & 1.0 \\
    \midrule 			

    & & \multicolumn{4}{c}{}                   \\ 
    
    & \makecell{$1 \times 10^{-6}$ \\ \\ } & 
    
\begin{tikzpicture}[every node/.style={minimum size=.625cm-\pgflinewidth, outer sep=0pt}]
    \draw[step=0.625cm,color=black] (0,0) grid (2.5,2.5);
    \node[fill=gray!41.93] at (0.3125,1.5625) {\tiny64.7};
    \node[fill=gray!27.46] at (0.3125,0.9375) {\tiny61.6};
    \node[fill=gray!27.09] at (0.3125,0.3125) {\tiny61.5};
    
    \node[fill=gray!60.6] at (0.9375,1.5625) {\tiny68.7};
    \node[fill=gray!51.92] at (0.9375,0.9375) {\tiny66.8};
    \node[fill=gray!27.37] at (0.9375,0.3125) {\tiny61.6};

    \node[fill=gray!54.44] at (1.5625,0.3125) {\tiny67.3};
    
    \node at (-0.625,+2.2) {\tiny$k$ = 20};
    \node at (-0.625,+1.575) {\tiny$k$ = 100};
    \node at (-0.625,+0.95) {\tiny$k$ = 200};
    \node at (-0.625,+0.325) {\tiny$k$ = 500};

    \node at (0.325,+2.75) {\begin{rotate}{60} {\tiny $\beta$ = 0.1} \end{rotate}};
    \node at (0.95,+2.75) {\begin{rotate}{60} {\tiny $\beta$ = 0.3} \end{rotate}};
    \node at (1.575,+2.75) {\begin{rotate}{60} {\tiny $\beta$ = 0.6} \end{rotate}};
    \node at (2.2,+2.75) {\begin{rotate}{60} {\tiny $\beta$ = 0.9} \end{rotate}};
\end{tikzpicture}

    & 

\begin{tikzpicture}[every node/.style={minimum size=.625cm-\pgflinewidth, outer sep=0pt}]
    \draw[step=0.625cm,color=black] (0,0) grid (2.5,2.5);
    \node[fill=gray!60.6] at (0.3125,2.1875) {\tiny68.3};
    \node[fill=gray!41.93] at (0.3125,1.5625) {\tiny64.7};
    \node[fill=gray!27.46] at (0.3125,0.9375) {\tiny61.6};
    \node[fill=gray!27.09] at (0.3125,0.3125) {\tiny61.5};
    
    \node[fill=gray!60.6] at (0.9375,1.5625) {\tiny68.7};
    \node[fill=gray!51.92] at (0.9375,0.9375) {\tiny66.8};
    \node[fill=gray!27.37] at (0.9375,0.3125) {\tiny61.6};

    \node[fill=gray!60.6] at (1.5625,0.9375) {\tiny68.6};
    \node[fill=gray!54.44] at (1.5625,0.3125) {\tiny67.3};
    
    \node at (0.325,+2.75) {\begin{rotate}{60} {\tiny $\beta$ = 0.1} \end{rotate}};
    \node at (0.95,+2.75) {\begin{rotate}{60} {\tiny $\beta$ = 0.3} \end{rotate}};
    \node at (1.575,+2.75) {\begin{rotate}{60} {\tiny $\beta$ = 0.6} \end{rotate}};
    \node at (2.2,+2.75) {\begin{rotate}{60} {\tiny $\beta$ = 0.9} \end{rotate}};
\end{tikzpicture}

     & 

\begin{tikzpicture}[every node/.style={minimum size=.625cm-\pgflinewidth, outer sep=0pt}]
    \draw[step=0.625cm,color=black] (0,0) grid (2.5,2.5);
    \node[fill=gray!60.6] at (0.3125,2.1875) {\tiny68.3};
    \node[fill=gray!41.93] at (0.3125,1.5625) {\tiny64.7};
    \node[fill=gray!27.46] at (0.3125,0.9375) {\tiny61.6};
    \node[fill=gray!27.09] at (0.3125,0.3125) {\tiny61.5};
    
    \node[fill=gray!60.6] at (0.9375,1.5625) {\tiny68.7};
    \node[fill=gray!51.92] at (0.9375,0.9375) {\tiny66.8};
    \node[fill=gray!27.37] at (0.9375,0.3125) {\tiny61.6};

    \node[fill=gray!76.24] at (1.5625,1.5625) {\tiny69.8};
    \node[fill=gray!60.6] at (1.5625,0.9375) {\tiny68.6};
    \node[fill=gray!54.44] at (1.5625,0.3125) {\tiny67.3};
    
    \node at (0.325,+2.75) {\begin{rotate}{60} {\tiny $\beta$ = 0.1} \end{rotate}};
    \node at (0.95,+2.75) {\begin{rotate}{60} {\tiny $\beta$ = 0.3} \end{rotate}};
    \node at (1.575,+2.75) {\begin{rotate}{60} {\tiny $\beta$ = 0.6} \end{rotate}};
    \node at (2.2,+2.75) {\begin{rotate}{60} {\tiny $\beta$ = 0.9} \end{rotate}};
\end{tikzpicture}

     &

\begin{tikzpicture}[every node/.style={minimum size=.625cm-\pgflinewidth, outer sep=0pt}]
    \draw[step=0.625cm,color=black] (0,0) grid (2.5,2.5);
    \node[fill=gray!60.6] at (0.3125,2.1875) {\tiny68.3};
    \node[fill=gray!41.93] at (0.3125,1.5625) {\tiny64.7};
    \node[fill=gray!27.46] at (0.3125,0.9375) {\tiny61.6};
    \node[fill=gray!27.09] at (0.3125,0.3125) {\tiny61.5};

    \node[fill=gray!80] at (0.9375,2.1875) {\tiny70.3};
    \node[fill=gray!60.6] at (0.9375,1.5625) {\tiny68.7};
    \node[fill=gray!51.92] at (0.9375,0.9375) {\tiny66.8};
    \node[fill=gray!27.37] at (0.9375,0.3125) {\tiny61.6};

    \node[fill=gray!76.24] at (1.5625,1.5625) {\tiny69.8};
    \node[fill=gray!60.6] at (1.5625,0.9375) {\tiny68.6};
    \node[fill=gray!54.44] at (1.5625,0.3125) {\tiny67.3};

    \node[fill=gray!56] at (2.1875,0.3125) {\tiny67.6};
    
    \node at (0.325,+2.75) {\begin{rotate}{60} {\tiny $\beta$ = 0.1} \end{rotate}};
    \node at (0.95,+2.75) {\begin{rotate}{60} {\tiny $\beta$ = 0.3} \end{rotate}};
    \node at (1.575,+2.75) {\begin{rotate}{60} {\tiny $\beta$ = 0.6} \end{rotate}};
    \node at (2.2,+2.75) {\begin{rotate}{60} {\tiny $\beta$ = 0.9} \end{rotate}};
\end{tikzpicture}

    \\

    & \makecell{$1 \times 10^{-9}$ \\ \\ } & 
    
\begin{tikzpicture}[every node/.style={minimum size=.625cm-\pgflinewidth, outer sep=0pt}]
    \draw[step=0.625cm,color=black] (0,0) grid (2.5,2.5);
    \node[fill=gray!41.93] at (0.3125,1.5625) {\tiny64.7};
    \node[fill=gray!27.46] at (0.3125,0.9375) {\tiny61.6};
    \node[fill=gray!27.09] at (0.3125,0.3125) {\tiny61.5};
    
    \node[fill=gray!51.92] at (0.9375,0.9375) {\tiny66.8};
    \node[fill=gray!27.37] at (0.9375,0.3125) {\tiny61.6};
    
    \node at (-0.625,+2.2) {\tiny$k$ = 20};
    \node at (-0.625,+1.575) {\tiny$k$ = 100};
    \node at (-0.625,+0.95) {\tiny$k$ = 200};
    \node at (-0.625,+0.325) {\tiny$k$ = 500};
\end{tikzpicture}

    & 

\begin{tikzpicture}[every node/.style={minimum size=.625cm-\pgflinewidth, outer sep=0pt}]
    \draw[step=0.625cm,color=black] (0,0) grid (2.5,2.5);
    \node[fill=gray!60.6] at (0.3125,2.1875) {\tiny68.3};
    \node[fill=gray!41.93] at (0.3125,1.5625) {\tiny64.7};
    \node[fill=gray!27.46] at (0.3125,0.9375) {\tiny61.6};
    \node[fill=gray!27.09] at (0.3125,0.3125) {\tiny61.5};
    
    \node[fill=gray!60.6] at (0.9375,1.5625) {\tiny68.7};
    \node[fill=gray!51.92] at (0.9375,0.9375) {\tiny66.8};
    \node[fill=gray!27.37] at (0.9375,0.3125) {\tiny61.6};

    \node[fill=gray!54.44] at (1.5625,0.3125) {\tiny67.3};
    
\end{tikzpicture}
     & 

\begin{tikzpicture}[every node/.style={minimum size=.625cm-\pgflinewidth, outer sep=0pt}]
    \draw[step=0.625cm,color=black] (0,0) grid (2.5,2.5);
    \node[fill=gray!60.6] at (0.3125,2.1875) {\tiny68.3};
    \node[fill=gray!41.93] at (0.3125,1.5625) {\tiny64.7};
    \node[fill=gray!27.46] at (0.3125,0.9375) {\tiny61.6};
    \node[fill=gray!27.09] at (0.3125,0.3125) {\tiny61.5};
    
    \node[fill=gray!60.6] at (0.9375,1.5625) {\tiny68.7};
    \node[fill=gray!51.92] at (0.9375,0.9375) {\tiny66.8};
    \node[fill=gray!27.37] at (0.9375,0.3125) {\tiny61.6};

    \node[fill=gray!60.6] at (1.5625,0.9375) {\tiny68.6};
    \node[fill=gray!54.44] at (1.5625,0.3125) {\tiny67.3};
\end{tikzpicture}

     &

\begin{tikzpicture}[every node/.style={minimum size=.625cm-\pgflinewidth, outer sep=0pt}]
    \draw[step=0.625cm,color=black] (0,0) grid (2.5,2.5);
    \node[fill=gray!60.6] at (0.3125,2.1875) {\tiny68.3};
    \node[fill=gray!41.93] at (0.3125,1.5625) {\tiny64.7};
    \node[fill=gray!27.46] at (0.3125,0.9375) {\tiny61.6};
    \node[fill=gray!27.09] at (0.3125,0.3125) {\tiny61.5};

    \node[fill=gray!60.6] at (0.9375,1.5625) {\tiny68.7};
    \node[fill=gray!51.92] at (0.9375,0.9375) {\tiny66.8};
    \node[fill=gray!27.37] at (0.9375,0.3125) {\tiny61.6};

    \node[fill=gray!76.24] at (1.5625,1.5625) {\tiny69.8};
    \node[fill=gray!60.6] at (1.5625,0.9375) {\tiny68.6};
    \node[fill=gray!54.44] at (1.5625,0.3125) {\tiny67.3};

\end{tikzpicture}

     \\

    & \makecell{$1 \times 10^{-12}$ \\ \\} & 
    
\begin{tikzpicture}[every node/.style={minimum size=.625cm-\pgflinewidth, outer sep=0pt}]
    \draw[step=0.625cm,color=black] (0,0) grid (2.5,2.5);
    \node[fill=gray!41.93] at (0.3125,1.5625) {\tiny64.7};
    \node[fill=gray!27.46] at (0.3125,0.9375) {\tiny61.6};
    \node[fill=gray!27.09] at (0.3125,0.3125) {\tiny61.5};
    
    \node[fill=gray!27.37] at (0.9375,0.3125) {\tiny61.6};
    
    \node at (-0.625,+2.2) {\tiny$k$ = 20};
    \node at (-0.625,+1.575) {\tiny$k$ = 100};
    \node at (-0.625,+0.95) {\tiny$k$ = 200};
    \node at (-0.625,+0.325) {\tiny$k$ = 500};
\end{tikzpicture}

    & 

\begin{tikzpicture}[every node/.style={minimum size=.625cm-\pgflinewidth, outer sep=0pt}]
    \draw[step=0.625cm,color=black] (0,0) grid (2.5,2.5);
    \node[fill=gray!41.93] at (0.3125,1.5625) {\tiny64.7};
    \node[fill=gray!27.46] at (0.3125,0.9375) {\tiny61.6};
    \node[fill=gray!27.09] at (0.3125,0.3125) {\tiny61.5};
    
    \node[fill=gray!60.6] at (0.9375,1.5625) {\tiny68.7};
    \node[fill=gray!51.92] at (0.9375,0.9375) {\tiny66.8};
    \node[fill=gray!27.37] at (0.9375,0.3125) {\tiny61.6};

    \node[fill=gray!54.44] at (1.5625,0.3125) {\tiny67.3};
    
\end{tikzpicture}

     & 

\begin{tikzpicture}[every node/.style={minimum size=.625cm-\pgflinewidth, outer sep=0pt}]
    \draw[step=0.625cm,color=black] (0,0) grid (2.5,2.5);
    \node[fill=gray!60.6] at (0.3125,2.1875) {\tiny68.3};
    \node[fill=gray!41.93] at (0.3125,1.5625) {\tiny64.7};
    \node[fill=gray!27.46] at (0.3125,0.9375) {\tiny61.6};
    \node[fill=gray!27.09] at (0.3125,0.3125) {\tiny61.5};
    
    \node[fill=gray!60.6] at (0.9375,1.5625) {\tiny68.7};
    \node[fill=gray!51.92] at (0.9375,0.9375) {\tiny66.8};
    \node[fill=gray!27.37] at (0.9375,0.3125) {\tiny61.6};

    \node[fill=gray!60.6] at (1.5625,0.9375) {\tiny68.6};
    \node[fill=gray!54.44] at (1.5625,0.3125) {\tiny67.3};
\end{tikzpicture}

     &

\begin{tikzpicture}[every node/.style={minimum size=.625cm-\pgflinewidth, outer sep=0pt}]
    \draw[step=0.625cm,color=black] (0,0) grid (2.5,2.5);
    \node[fill=gray!60.6] at (0.3125,2.1875) {\tiny68.3};
    \node[fill=gray!41.93] at (0.3125,1.5625) {\tiny64.7};
    \node[fill=gray!27.46] at (0.3125,0.9375) {\tiny61.6};
    \node[fill=gray!27.09] at (0.3125,0.3125) {\tiny61.5};

    \node[fill=gray!60.6] at (0.9375,1.5625) {\tiny68.7};
    \node[fill=gray!51.92] at (0.9375,0.9375) {\tiny66.8};
    \node[fill=gray!27.37] at (0.9375,0.3125) {\tiny61.6};

    \node[fill=gray!60.6] at (1.5625,0.9375) {\tiny68.6};
    \node[fill=gray!54.44] at (1.5625,0.3125) {\tiny67.3};
\end{tikzpicture}

     \\

    & \makecell{$1 \times 10^{-15}$ \\ \\ } & 
    
\begin{tikzpicture}[every node/.style={minimum size=.625cm-\pgflinewidth, outer sep=0pt}]
    \draw[step=0.625cm,color=black] (0,0) grid (2.5,2.5);
    \node[fill=gray!41.93] at (0.3125,1.5625) {\tiny64.7};
    \node[fill=gray!27.46] at (0.3125,0.9375) {\tiny61.6};
    \node[fill=gray!27.09] at (0.3125,0.3125) {\tiny61.5};
    
    \node[fill=gray!27.37] at (0.9375,0.3125) {\tiny61.6};
    
    \node at (-0.625,+2.2) {\tiny$k$ = 20};
    \node at (-0.625,+1.575) {\tiny$k$ = 100};
    \node at (-0.625,+0.95) {\tiny$k$ = 200};
    \node at (-0.625,+0.325) {\tiny$k$ = 500};
\end{tikzpicture}

    & 

\begin{tikzpicture}[every node/.style={minimum size=.625cm-\pgflinewidth, outer sep=0pt}]
    \draw[step=0.625cm,color=black] (0,0) grid (2.5,2.5);
    \node[fill=gray!41.93] at (0.3125,1.5625) {\tiny64.7};
    \node[fill=gray!27.46] at (0.3125,0.9375) {\tiny61.6};
    \node[fill=gray!27.09] at (0.3125,0.3125) {\tiny61.5};
    
    \node[fill=gray!51.92] at (0.9375,0.9375) {\tiny66.8};
    \node[fill=gray!27.37] at (0.9375,0.3125) {\tiny61.6};

    \node[fill=gray!54.44] at (1.5625,0.3125) {\tiny67.3};
\end{tikzpicture}
     & 

\begin{tikzpicture}[every node/.style={minimum size=.625cm-\pgflinewidth, outer sep=0pt}]
    \draw[step=0.625cm,color=black] (0,0) grid (2.5,2.5);
    \node[fill=gray!41.93] at (0.3125,1.5625) {\tiny64.7};
    \node[fill=gray!27.46] at (0.3125,0.9375) {\tiny61.6};
    \node[fill=gray!27.09] at (0.3125,0.3125) {\tiny61.5};
    
    \node[fill=gray!60.6] at (0.9375,1.5625) {\tiny68.7};
    \node[fill=gray!51.92] at (0.9375,0.9375) {\tiny66.8};
    \node[fill=gray!27.37] at (0.9375,0.3125) {\tiny61.6};

    \node[fill=gray!54.44] at (1.5625,0.3125) {\tiny67.3};
\end{tikzpicture}

     &

\begin{tikzpicture}[every node/.style={minimum size=.625cm-\pgflinewidth, outer sep=0pt}]
    \draw[step=0.625cm,color=black] (0,0) grid (2.5,2.5);
    \node[fill=gray!41.93] at (0.3125,1.5625) {\tiny64.7};
    \node[fill=gray!27.46] at (0.3125,0.9375) {\tiny61.6};
    \node[fill=gray!27.09] at (0.3125,0.3125) {\tiny61.5};

    \node[fill=gray!60.6] at (0.9375,1.5625) {\tiny68.7};
    \node[fill=gray!51.92] at (0.9375,0.9375) {\tiny66.8};
    \node[fill=gray!27.37] at (0.9375,0.3125) {\tiny61.6};

    \node[fill=gray!60.6] at (1.5625,0.9375) {\tiny68.6};
    \node[fill=gray!54.44] at (1.5625,0.3125) {\tiny67.3};
\end{tikzpicture}

     \\

    \bottomrule
  \end{tabular}
\end{table*}

\bibliographystyle{aaai}
\bibliography{aaai}

\end{document}